\documentclass[letterpaper, 10 pt, conference]{ieeeconf}  
\usepackage{verbatim}
\usepackage{graphicx}
\usepackage{float}
\usepackage{booktabs}
\usepackage{amsmath,amssymb,amsfonts}
\usepackage{mathtools}
\usepackage{bm}
\usepackage{array}   
\usepackage{hyperref}
\usepackage{url}

\IEEEoverridecommandlockouts                              

\usepackage[dvipsnames,table]{xcolor}

\title{\LARGE \bf
From Score to Sound: An End-to-End MIDI-to-Motion Pipeline for Robotic Cello Performance
}

\author{Samantha Sudhoff, Pranesh Velmurugan, Jiashu Liu, Vincent Zhao, Yung-Hsiang Lu, Kristen Yeon-Ji Yun} 

\begin{document}

\maketitle
\thispagestyle{empty}
\pagestyle{empty}

\begin{abstract}

 Robot musicians require precise control to obtain proper note accuracy, sound quality, and musical expression. Performance of string instruments, such as violin and cello, presents a significant challenge due to the precise control required over bow angle and pressure to produce the desired sound. While prior robotic cellists focus on accurate bowing trajectories, these works often rely on expensive motion capture techniques, and fail to sightread music in a human-like way. 
 We propose a novel \textbf{end-to-end MIDI score to robotic motion pipeline} which converts musical input directly into collision-aware bowing motions for a UR5e robot cellist. Through use of Universal Robot Freedrive feature, our robotic musician can achieve human-like sound without the need for motion capture. Additionally, this work records live joint data via Real-Time Data Exchange (RTDE) as the robot plays, providing labeled robotic playing data from a collection of five standard pieces to the research community. To demonstrate the effectiveness of our method in comparison to human performers, we introduce the \textbf{Musical Turing Test}, in which a collection of 132 human participants evaluate our robot's performance against a human baseline. Human reference recordings are also released, enabling direct comparison for future studies. This evaluation technique establishes the first benchmark for robotic cello performance. Finally, we outline a residual reinforcement learning methodology to improve upon baseline robotic controls, highlighting future opportunities for improved string-crossing efficiency and sound quality. 
\\


\end{abstract}

\section{INTRODUCTION}

Robotic musicianship is an interdisciplinary topic encompassing robot control, mechanics, music, human-robot interaction, and more. As such, various robot musicians are focused on different goals. While some robot musician projects are developed as an exploration of sound quality or expression, others are designed to ensure the highest and fastest note accuracy possible. String instruments with bows such as violin and  cello present especially difficult challenges, as musical quality depends not only on correct pitch and timing, but also on nuanced control of bowing position, direction, and pressure. 

Prior approaches to robotic musicianship of bowed instruments have relied heavily on motion capture systems or pre-recorded demonstrations to generate expressive trajectories ~\cite{pemies-bowing-framework, pemies-cello-kinematics, park-sound-quality}. While effective, these approaches are costly, time-intensive, and often limited to replaying a fixed repertoire rather than generalizing to new musical input. Furthermore, many systems emphasize mechanical accuracy but fail to analyze human input and enjoyment. 

\begin{figure}[h]
    \centering
    \includegraphics[width=0.9\linewidth]{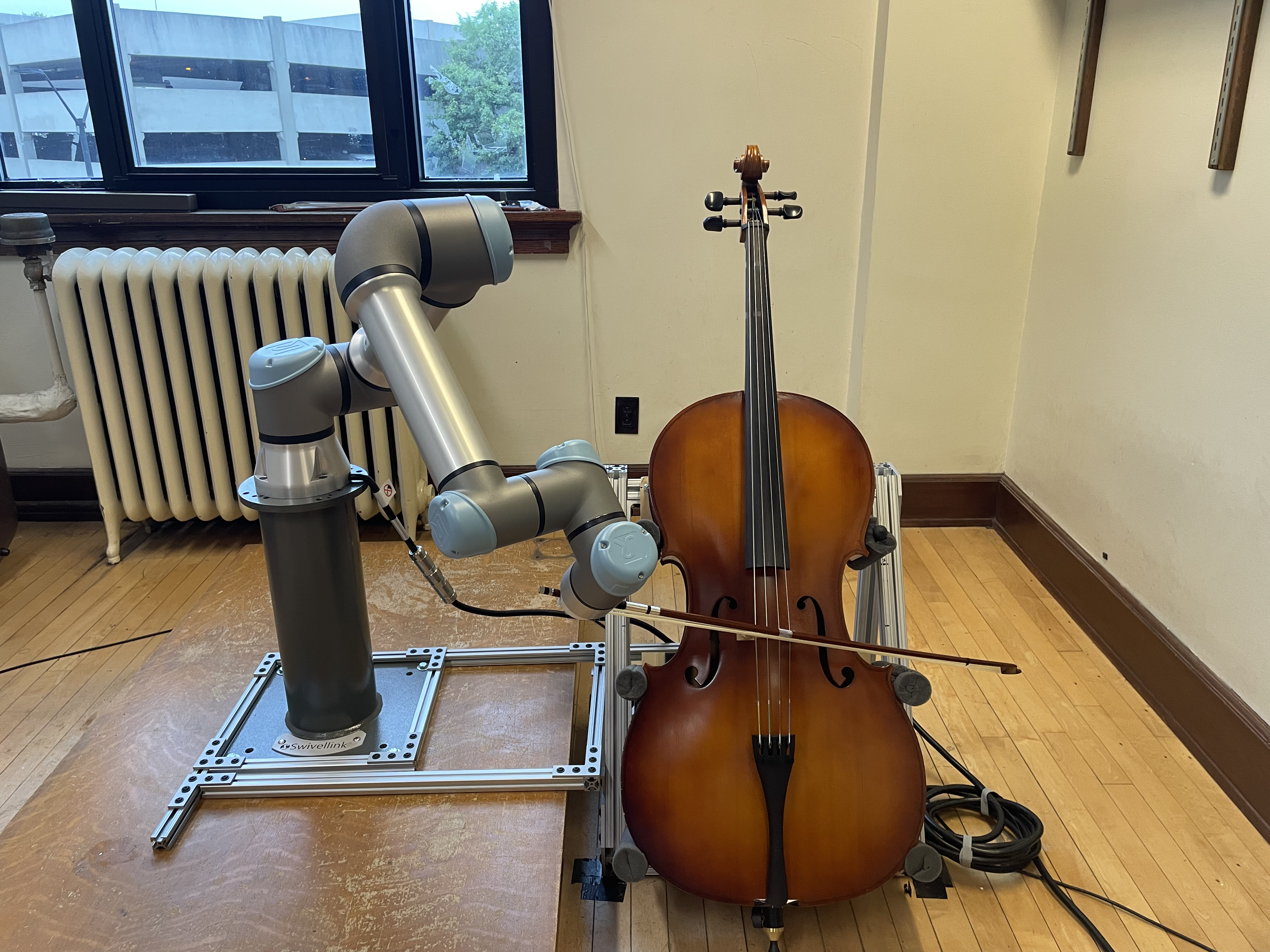}
    \caption{Experimental setup of the robotic cellist, including the UR5e robotic arm, cello, custom stand, and metal frame.}
    \label{figure:real-life-setup}
\end{figure}

In our work, we propose a novel end-to-end MIDI-to-motion pipeline for robotic cello performance. Our system converts symbolic input in the form of a MIDI score directly into executable robotic trajectories for a UR5e collaborative robot arm ~\cite{universalrobots-ur5e}. This approach does not require motion capture or human demonstrations, enabling the robot to autonomously sight-read and perform previously unseen musical pieces. The pipeline integrates musical score parsing, trajectory generation, and robot execution in a unified framework, allowing for effective open-string bowed performances of beginner cello pieces.

To evaluate the effectiveness of our system, we introduce a Musical Turing Test. In this study, human participants are presented with paired recordings of the same musical piece, one performed by our robotic cellist and the other by a human cellist. Listeners are tasked with identifying which performance is the robot. By analyzing the results across five standard beginner cello pieces, we assess whether our robot’s playing is distinguishable from that of an intermediate human performer. We find that for non-musicians ($n=29$), average accuracy in determining robotic versus human performer is $55.88\%$. Because each audio clip requires a binary choice (robot versus human), the chance level expectation is $50\%$. The observed accuracy, just $5.88\%$ higher than the random-guess baseline, indicates that participants are effectively guessing, and thus unable to distinguish between the robotic performer and the human musician. 

Our contributions are as follows:
\begin{itemize}
    \item We present an end-to-end pipeline that converts MIDI input directly into robotic motions for cello performance, eliminating the need for motion capture.
    \item We create a dataset of synchronized robot trajectories, audio, and video recordings of both robot and human performances to support reproducibility and future research. A demonstration video of our robot's performance will be available as a complementary submission. 
    \item We evaluate our system through a human-subject Musical Turing Test, providing insights into how listeners perceive robotic cello performance relative to human playing.
\end{itemize}

Through these contributions, we demonstrate that an end-to-end MIDI-to-motion approach can achieve musically convincing robotic performance, offering a promising foundation for future research toward human-level robotic technology.

\section{RELATED WORKS}
\subsection{Robotic Musicianship}

Robotic musicianship involves many areas such as musical perception, control, and human–robot interaction. As shown in \textit{Robotic Musicianship: Embodied Artificial Creativity and Musical Interaction} \cite{bretan-weinberg-survey,cmt-book}, robotic performance is  a technical task as well as a social and artistic one that depends on perception, timing, and gesture. \cite{bretan-weinberg-survey}. Early systems such as the MUBOT project in Japan explore recorder, violin, and cello robots as an extension of the human performer, emphasizing both the technical and cultural dimensions of musician robots \cite{kajitani-mubot}. Other systems include flute-playing robots which use neural networks to model vibrato and note-duration variations from professional players, showing how deviations from the score are critical to musicality \cite{solis-flute-nn}. These works highlight the difficulty of blending musicality with technical accuracy, a challenge that becomes even greater for string instruments. Producing a stable and expressive tone on such instruments requires continuous control of bow motion and precise management of the contact point on the string.

\subsection{Bowed-String and Cello Robots}
Bowed string instruments impose tight constraints on bow–string interaction, as tone quality depends heavily on bow force, velocity, and sounding point. Robotic violin systems have addressed these factors through damping mechanisms, anthropomorphic robotic fingering mechanisms, and score-informed control strategies \cite{min-violin-damper,park-violin-sq,horigome-violin-motion}. Such approaches emphasize the difficulty of stabilizing bow–string interaction while still producing rich and expressive sound. To achieve high perceived levels of expressiveness and reach various musical timbres, tempo and sound-pressure variations are an effective strategy of producing both bright and dark musical tones \cite{shibuya-violin-timbre}.

Robotic cello systems remain comparatively rare. Lampis \emph{et al.} focus on a singular cello string, examining bridge force components as the string is played by a UR5e robotic arm \cite{lampis-bridge-force-2023}. This work is extended to development of a monochord robotic arm for bowing and plucking excitation, allowing for controlled bowing parameters ~\cite{mayer-monochord-2024}. Through use of motion capture techniques, P\`amies-Vil\`a \emph{et al.} replicate expert bowing kinematics, producing detailed velocity and acceleration profiles but remaining limited to dependencies on human demonstration data \cite{pemies-cello-kinematics,pemies-bowing-framework}. These studies advance acoustic understanding of cello performance, but cannot generalize to unseen musical scores. In contrast, our system converts symbolic input (MIDI) directly into bowing motions across all strings, avoiding reliance on motion capture and enabling direct execution from score to performance.

To our knowledge, no prior robotic cello system has converted symbolic input such as a MIDI file directly into robotic bowing trajectories. While prior robotic musicians have used MIDI for percussive or keyboard instruments such as marimbas \cite{weinberg-marimba} or guitar \cite{guitarbot-lemur-2003}, bowed-string applications remain underexplored. Our system advances this direction by incorporating bow direction, articulation, and timing directly from symbolic score data, producing physically executable trajectories on a full-size cello. Furthermore, prior cello works rely primarily on mimicking human cellist performance through expensive motion capture techniques, where focus is on kinematic similarity rather than on auditory outcomes. By contrast, our work does not require such methods, shifting the focus from geometric replication to perceptual measures of sound quality and musicality. This is done through a Musical Turing Test, directly aligning evaluation with the goals of expressive performance. A direct comparison of our robotic bowed musician to prior methods can be found in Table \ref{tab:bowed-comparison}. 

\subsection{Human-Listener Evaluation Methods}
Listener-based evaluation has long been used to assess whether algorithmic or robotic performances are perceived as “human-like.” In generative music research, Musical Turing Test paradigms ask listeners to distinguish human from machine or to rate human-likeness under blind conditions \cite{roda-musical-tt}. Robotic-music user studies additionally examine audience perception and interaction quality in live or lab settings \cite{beatbots-user-study}. Our evaluation follows this tradition by presenting paired human/robot recordings of the same pieces and measuring discriminability at the listener level.

\begin{table*}[ht]
\centering
\caption{Comparison of robotic bowed-string instrument methods.}
\label{tab:bowed-comparison}
\renewcommand{\arraystretch}{1.5} 
\begin{tabular}{p{2cm} p{1.2cm} p{1.4cm} p{3cm} p{1.6cm} p{5.5cm}}
\toprule
\textbf{Work} & \textbf{Instrument} & \textbf{Scope} & \textbf{Dependency on Human Data} & \textbf{Generalization to new pieces} & \textbf{Evaluation Paradigm} \\
\midrule
K Horigome et al. \cite{horigome-violin-motion} &
Violin &
Four strings &
Moderate (requires human data for sound-pressure targets) &
Limited &
Simulation-based learning success; comparison of sound pressure to targets \\
\midrule
W Jo et al. \cite{park-violin-sq} &
Violin &
One string &
Moderate (imitates human technique) &
None &
Acoustic analysis + expert evaluation \\
\midrule
K Shibuya et al. \cite{shibuya-violin-timbre} &
Violin &
Four strings &
Moderate (uses deviations modeled from human recordings) &
Limited &
Human-listener study (n=10) judging timbre (bright/dark) \\
\midrule
A Mayer et al. \cite{mayer-monochord-2024} &
Cello &
One string &
Low &
None &
Acoustic and force/vibration measurements \\
\midrule
M Pàmies-Vilà et al. \cite{pemies-cello-kinematics,pemies-bowing-framework} &
Cello &
G, D, A strings &
High (requires expert motion capture) &
Limited &
Kinematic (position/velocity) and acoustic error (contact, RMS, spectral) \\
\midrule
\textbf{Our method} &
Cello &
Four strings &
Low &
High &
Human-listener study (n=132) Musical Turing Test (human vs. robot) \\
\bottomrule
\end{tabular}
\end{table*}

\begin{figure}[H]
    \centering
    \includegraphics[width=1.0\linewidth]{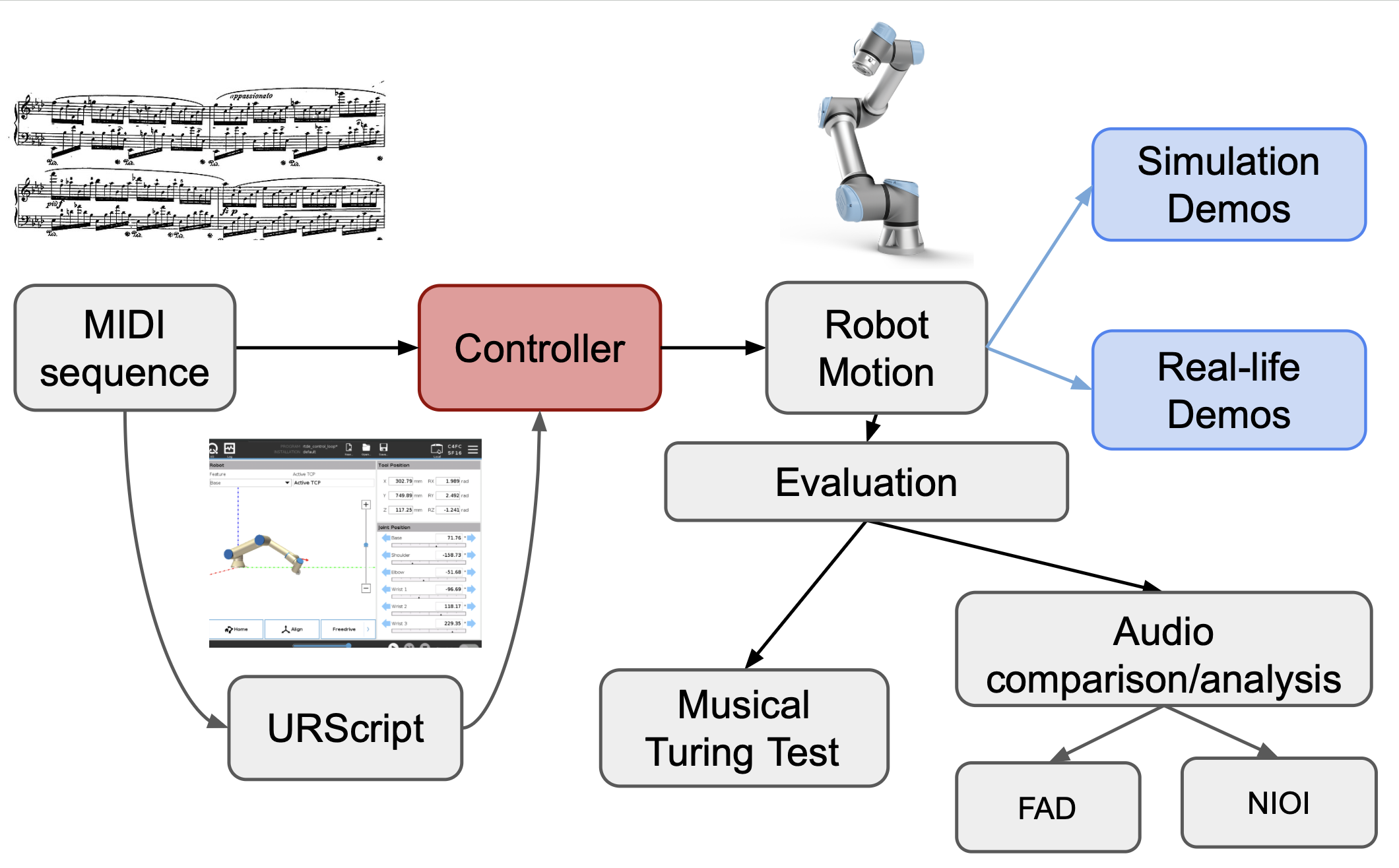}
    \caption{MIDI-to-Motion pipeline which converts a musical sequence into robotic motions, allowing for capture of demonstration data.}
    \label{figure:flowchart}
\end{figure}

\section{METHODOLOGY}
     Figure \ref{figure:flowchart} presents the end-to-end MIDI score–to–robotic motion pipeline, which converts cello scores into robotic joint position trajectories over time. This pipeline effectively allows a robotic arm to automatically play open-string bowings of cello pieces, given the MIDI-formatted score. 

\subsection{Expert-Defined Bowing Primitives}
    To guide the bowing motions for each of the four cello strings (A, D, G, C), a student musician from our research lab predefined the frog (start) and tip (end) Cartesian base positions [x, y, z, rx, ry, rz] in meters and radians  define the linear path of the bow using the robotic UR5e Freedrive feature. For a given string, rotation of the tool position is kept constant from frog to tip. Maintenance of a fixed rotational parameter value per-string allows for simple calculations of string crossing motions and to easy definition of fractional positions across the bow for each note. Eight total waypoints are hand-defined at points of flat bow-string contact, with two waypoints per cello string. To verify proper contact point with the bow for these collected waypoints, bow position was fine-tuned from various angles. For testing, the robot was jogged from frog waypoint to tip waypoint, where a human cellist from the research lab verified and adjusted the positions accordingly to produce a consistent sound. The eight collected waypoints are used as primitives to use in defining the linear bow paths of the robot as it is given various notes. To account for string crossings (between-string motions), a sequence of three steps is followed: (i) the cello bow is moved out from the string plane for safety and collision prevention, (ii) the bow is translated to the target string at the same fractional position along that string's frog-tip line as the current position, and (iii) the bow is reseated to contact the target string at a defined point along its frog-tip line. Each out-position is hand-verified to determine collision prevention. 
\subsection{MIDI-to-Motion Translation Framework}
\begin{figure}[h]
    \centering
    \includegraphics[width=1.0\linewidth]{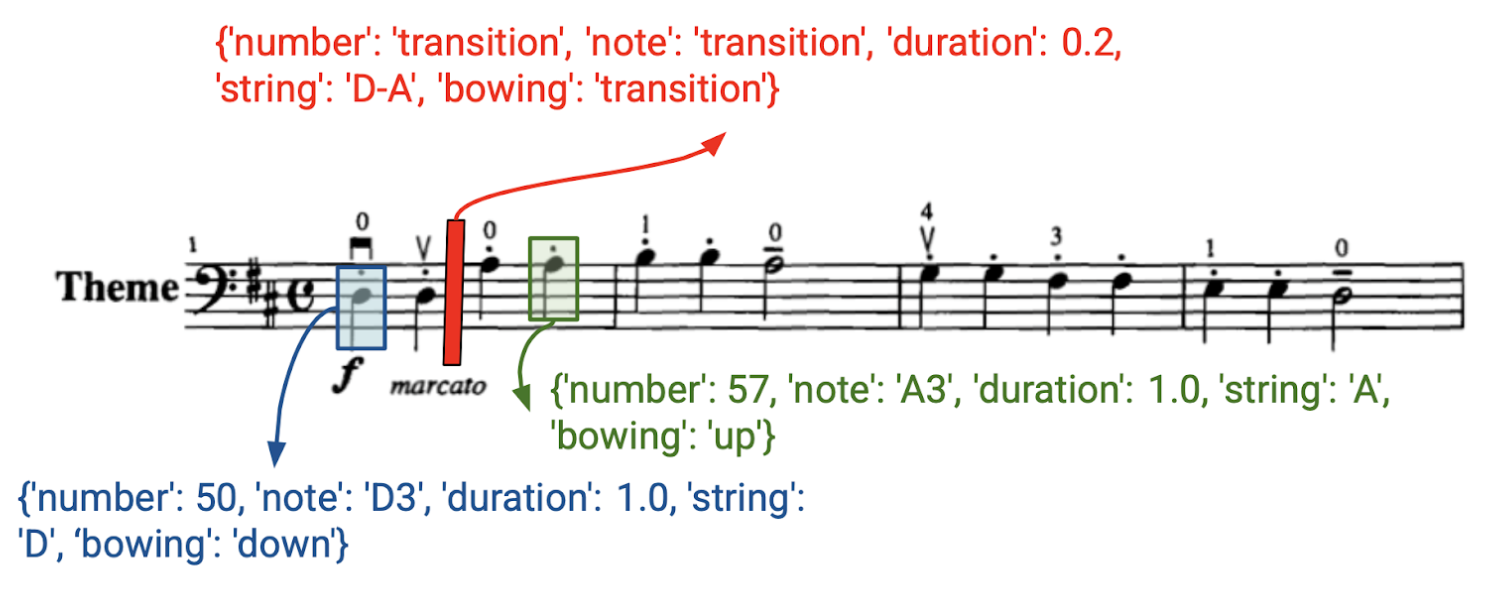}
    \caption{Method for translating a MIDI music score into a parameterized note sequence with attached information on note name, note duration, cello string, and bowing.}
    \label{fig:midi-motion-pipeline}
\end{figure} 

From the eight string primitives denoted $p_1, ... p_8$, each a 6-tuple $[x, y, z, rx, ry, rz]$, we develop a procedure to perform the specified note sequence from an input MIDI file. First, the MIDI file is converted into a sequence of notes by use of the Python Mido Library ~\cite{mido-github} as shown in Figure \ref{fig:midi-motion-pipeline}. The attached information for each note in the sequence can be represented in minimal form by $(s_i, d_i, b_i)$, where $s_i \in \{A, D, G, C\}$ is the target string, $d_i$ is the target duration in seconds, and $b_i \in \{\downarrow, \uparrow\}$ is the target bowing. Bowings for a given piece are assumed to alternate between down bow (frog-to-tip) and up bow (tip-to-frog) with each changing note, unless otherwise specified by an optional input file with provided target bowings. 

\begin{figure}[H]
    \centering
    \includegraphics[width=1.0\linewidth]{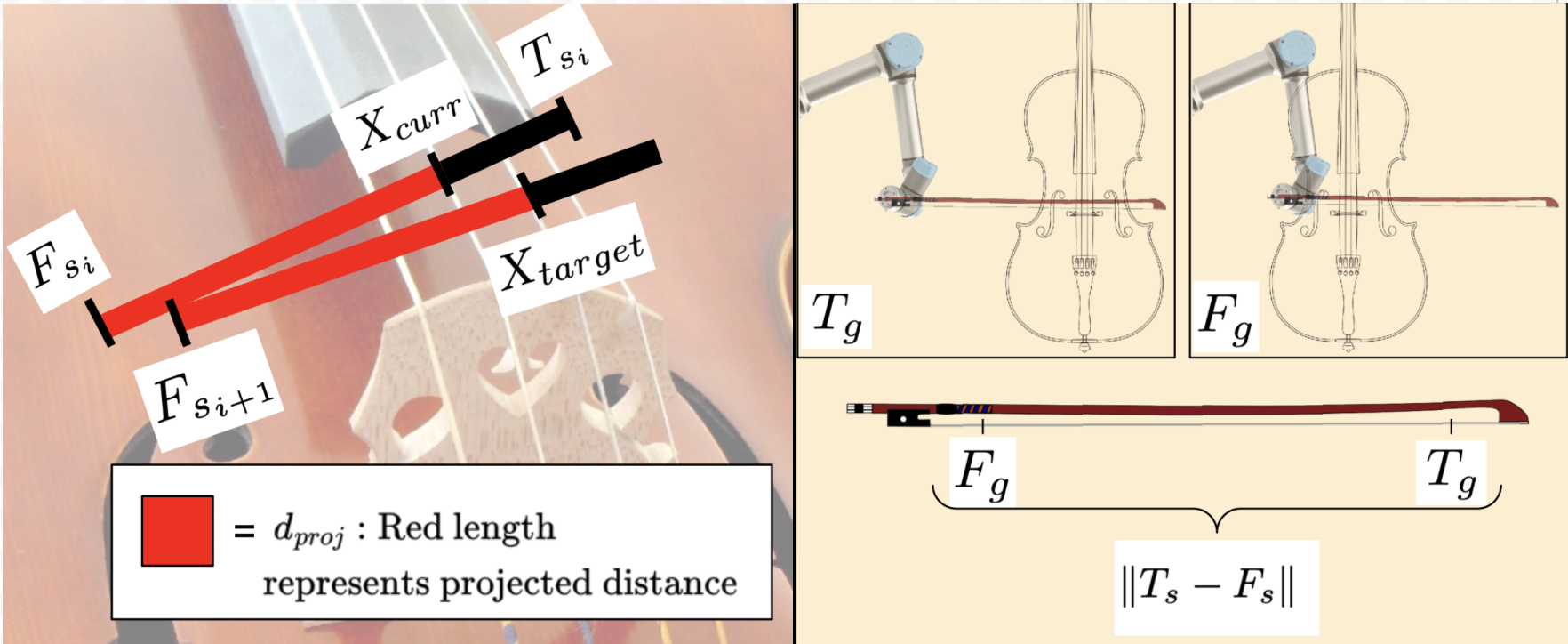}
    \caption{Illustration of robotic bowing primitives and string crossings.
Left: example of a string crossing movement, showing projection of the current bow fraction onto the target string. 
Right: example of frog ($F_s$) and tip ($T_s$) waypoints with the corresponding bow length $\lVert T_s - F_s \rVert$. Note: UR5e robot image is adapted from ~\cite{ur5e-image}.} 
    \label{fig:geometry}
\end{figure}

For each string $s$ in the generated note sequence, we refer to the hand-defined frog ($F_s$) and tip ($T_s$) primitives (as described in Section A. amd shown in Figure \ref{fig:geometry}) :

\[
a_{\text{bow\_poses}} = \left\{
\begin{aligned}
T_A &= 
\begin{bmatrix}
0.4342 \\ 0.3790 \\ 0.2700 \\ -1.5440 \\ -2.3550 \\ 1.3470
\end{bmatrix}, \\
F_A &= 
\begin{bmatrix}
0.3007 \\ 0.7936 \\ 0.0997 \\ -1.5440 \\ -2.3550 \\ 1.3470
\end{bmatrix}
\end{aligned}
\right\}, \quad \text{etc. for D, G, C.}
\]

The end-effector pose of the robotic arm is interpolated linearly along the bowing line using a fractional parameter $u \in [0,1]$:
\[
X_s(u,b) =
\begin{cases}
(1-u)F_s + uT_s, & b = \downarrow \; (\text{frog} \to \text{tip}) \\
(1-u)T_s + uF_s, & b = \uparrow \; (\text{tip} \to \text{frog})
\end{cases}
\]
with constant orientation $R_s$ for each string plane. 

The target displacement length along the bow required for a given note is aligned with note duration $d_i$. As such, the linear distance the bow must travel for a given note is represented by:

\[
L_{\text{target}} = \alpha(d_i)\,\lVert T_s - F_s \rVert
\]
where $L_{\text{target}}$ is the distance to travel, $\lVert T_s - F_s \rVert$ is the L2 distance from frog to tip for the current string, and $\alpha(d_i)$ is a cosine interpolation function which maps bow duration to bow stroke length, given by the following:
\[\alpha(d_i) = \frac{1-cos(\pi \cdot \frac{d}{3})}{2}, d \in [0, 3]\] 

The scaling function ensures that short duration bow motions are not choppy, and additionally caps the total stroke length at three seconds for long notes. From the calculated bow travel distance, direction, and time, the controller issues a linear motion command to traverse the bow fraction required in the specified amount of time.

If the available bow length from the current position is insufficient to reach $L_{\text{target}}$, there are three possible cases handled by the system:

\begin{enumerate}
    \item \textbf{Continue in the same direction:} In this case, the remaining bow displacement is less than the defined tolerance of $0.025\,\text{m}$ from the computed target, and  the stroke is terminated by moving to the corresponding endpoint pose ($F_s$ or $T_s$), completing the motion without additional correction.
    \item \textbf{Flip bowing direction:} Reverse direction $b_i$ and continue with the current pose.
    \item \textbf{Reset bow:} Reposition to the frog position ($F_s$) for down bow or tip ($T_s$) position for up bow, as it is not possible to reverse the bow for the target duration.
\end{enumerate}

In the case of a string crossing (i.e. where $s_{i+1} \neq s_i$), the controller executes a three-step transition to avoid unwanted string collisions and preserve bow alignment with the next-string line. First, the end-effector moves outward from the current TCP (tool center point) pose by a fixed offset vector $\Delta_{out}= [0.1, 0.0, 0.5, 0, 0, 0]$ to create a safe clearance for the bow to rotate towards the correct string. Then, given our string current string $s_i$'s fractional position from the frog, we project this distance onto the frog-to-tip line of our target string $s_{i+1}$. This bow-distance $d_{proj}$, as shown in Figure \ref{fig:geometry}, can be represented by: 
\[d_{proj}=\frac{\lvert\lvert X_{curr} - F_{s_i} \lvert \lvert}{\lvert \lvert T_{s_i} - F_{s_i} \lvert \lvert} \cdot \lvert \lvert T_{s_{i+1}} - F_{s_{i+1}} \lvert \lvert\] 
Where $X_{curr}$ is our current position, $\frac{\lvert\lvert X_{curr} - F_{s_i} \lvert \lvert}{\lvert \lvert T_{s_i} - F_{s_i} \lvert \lvert}$ is the current fraction along the bow, and $\lvert \lvert T_{s_{i+1}} - F_{s_{i+1}} \lvert \lvert$ is the length of the target string $s_{i+1}$. As such, we can easily calculate our target position $X_{target}$ as:
\[X_{target}=F_{s_{i+1}}+\frac{T_{s_{i+1}} - F_{s_{i+1}}}{\lvert\lvert T_{s_{i+1}} - F_{s_{i+1}}\lvert\lvert} \cdot d_{proj}\]

From the position $X_{curr}+\Delta_{out}$, we first make a joint movement towards $X_{target}+\Delta_{out}$, and then finally make a small movement to bring the end-effector down onto the target string at $X_{target}$. This three-step process ensures collision-free transitions to the target string at the proper fractional position along the bow.

\section{EXPERIMENTS}

\subsection{Robotic Arm}

Our experimental setup consists of a 6-DoF UR5e robotic arm by Universal Robot ~\cite{universalrobots-ur5e}. This robotic arm plays a full-size cello which is mounted to a custom stand and attached to the robotic arm with a metal frame, as shown in Figure~\ref{figure:real-life-setup}. This attachment ensures proper calibration between the instrument and the robot's coordinate system. The cello bow is attached to the end-effector of the UR5e using a custom 3D-printed gripper, designed  in our research lab. 

\subsection{Data Collection}
For evaluation of our system, a dataset is assembled from five standard beginner-level cello pieces. A description of these pieces can be found in Section~V.B.1 (Music Description). For each piece, we collect both human-performed and robot-performed audio recordings, and additionally collect robotic video recordings. Detailed robotic motion logs obtained using Real-Time Data Exchange (RTDE) interface of the UR5e robot is used, where positional and note information is gathered every $\sim$0.01 seconds. In addition, more highly detailed data is collected using Universal Robot's simulation software: URSim. All data collected for usage of this project will be available after
this paper is accepted.


The released dataset contains 88130 rows of URSim data, accounting for around 15 total minutes of robotic motion. Available fields for this dataset include : [timestamp-robot, time-elapsed-sec, event-flag, event-label, current-event-type, current-note-number, current-note-name, current-string, current-bowing, remaining-duration-sec, TCP-pose-x, TCP-pose-y, TCP-pose-z, TCP-pose-rx, TCP-pose-ry, TCP-pose-rz, q-base, q-shoulder, q-elbow, q-wrist1, q-wrist2, q-wrist3]. This dataset enables detailed analysis of robotic bowing mechanics, and sets up future usage in tasks such as Behavioral Cloning. 

\begin{figure*}[t]
    \centering
    \includegraphics[width=1.0\linewidth]{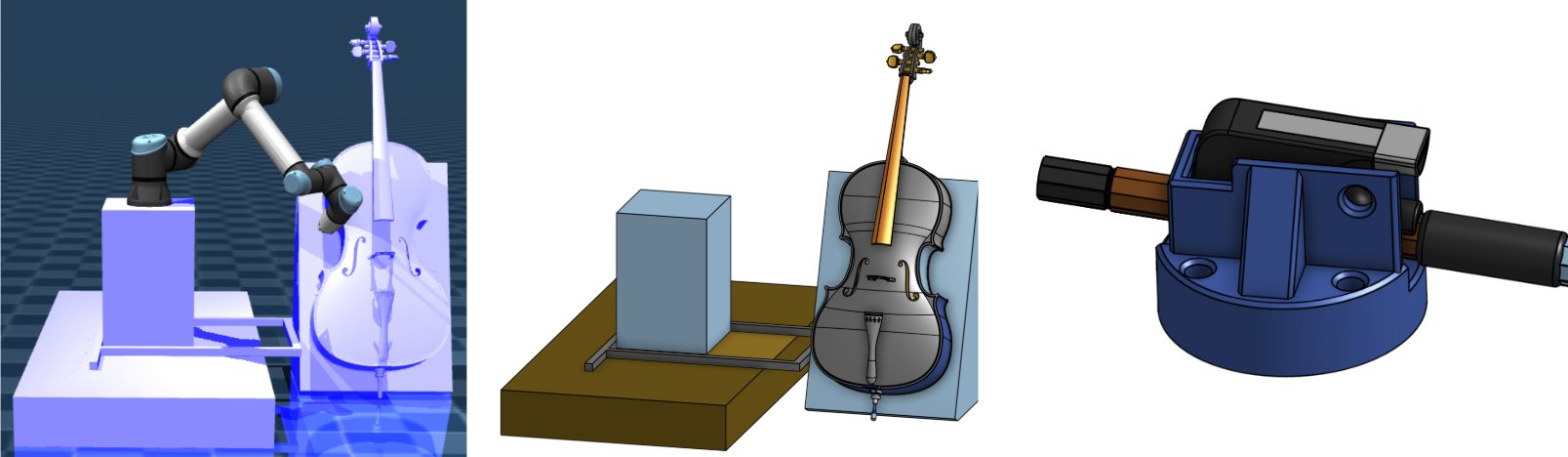}
    \caption{System setup for robotic cello playing. (Left) UR5e robot arm positioned with a cello in MuJoCo simulation, (Middle) CAD design of the instrument support and base platform, (Right) custom-designed bow gripper for precise bowing control.}
    \label{figure:robo_arm}
\end{figure*}

\subsection{Simulation Setup}
Project includes an environmental setup created in the MuJoCo simulation software. To accurately train a model transferable from simulation to real life, we created a MuJoCo environment closely matching the real-life lab setup. Measurements were taken of the physical setup, and a simplified recreation was subsequently made in Onshape using those measurements. A cello model matching the instrument used was obtained online [14].

The second step in ensuring the accuracy of the simulated environment is correctly modeling the bow placement. The bow gripper (an end effector that attaches the bow to the robot) was modeled in CAD using Onshape. A bow model closely matching the physical bow was obtained online [15].

The cello strings were represented as ellipsoids in MuJoCo, and the bow hair was similarly modeled as an ellipsoid. This approach allowed the simulation to detect contact as collisions between the bow hair and the cello strings. It also facilitated reinforcement learning by making it clearer when the robot successfully played a note. While RL is not currently the focus, the simulated environment still provides value. The environment allows for testing motion without the physical robot. This helps ensure that new motion are safe before they are attempted in the physical robot. Furthermore, in the future, when RL is used for optimization, the simulated environment will prove useful.

\section{Human-Subject Evaluation and Results}

We explore the effectiveness of our system through a Musical Turing Test: a human-subject evaluation where participants are tasked with comparing the performance of our robotic musician to recorded human performances of the same piece. The objective of this evaluation is to determine whether our robotic musician is distinguishable from that of an intermediate human cellist from our research lab (10 years playing experience). Specifically, this study seeks to explore the folllowing research question: \textit{Is a robotic cellist capable of comparable performance to that of an intermediate human musician?}. Additionally, through this study we explore human preferences in musical performance as well as explore potential biases in terms of robotic musicians. All questions used in the survey will be released to the research community alongside our recorded collection of human and robotic open string performances. 

\subsection{Study Participants}

Recruitment for the human-subject evaluation involved reaching out to
university students and faculty members. A total of 132 participants were recruited to participate in the study.:$68\%$ of participants were male, and $28\%$ were female; the others did not answer. Most subjects ($96\%$) were between 18-24. In regards to musical experience, the majority ($73\%$) of participants had some form of experience playing an instrument; $34\%$ of participants were either professional or student musicians; $33\%$ of participants played for leisure, and $5\%$ of participants were composers or other forms of musician. 

\subsection{Music Description}

A selection of five standard beginning pieces were selected for usage in this study, with the majority of these pieces found in Suzuki Cello School Volume 1 ~\cite{suzuki-volume-1}: (1) "Allegro", (2) "Perpetual Motion", (3) "Twinkle, Twinkle, Little Star Variations", by Shinichi Suzuki and (4) "Minuet No. 2" by Johann Sebastian Bach. Additionally, a piece has been taken from Suzuki Cello School Volume 2 ~\cite{suzuki-volume-2}: (5) "Long, Long Ago" by T. H. Bayley. The selection of music for this study is aimed at targeting various bow speeds, string crossings, and types of expression required. To evaluate effectiveness of our robotic cellist in perform with fast articulation, "Allegro" and "Perpetual Motion" provide useful explorations. To evaluate our robot's performance in slow and long bowing passages, "Long, Long Ago" is used. "Twinkle, Twinkle, Little Star Variations" provides a classic and standard piece which focuses on the A and D strings. Lastly, "Minuet No. 2" provides the greatest challenge: a lengthy piece of standard cello repertoire which requires complex rhythms and is of greater duration than the other four pieces. 

A human cellist from our research lab produced open-string recordings for all five pieces. Additionally, open-string recordings obtained via hand-converting the five musical scores to MIDI format and then recording robotic audio produced from the MIDI-to-Motion pipeline of our robot was gathered for each piece, where each piece was converted to a sequence of notes using our described methods and then played by a UR5e Robotic arm. The obtained robotic recordings for these pieces are trimmed to eliminate noisy string crossings; This is a needed standardization for evaluation of cello performance which is disclosed to the study subjects. This trimming effect is used eliminate potential non-performance related factors. String crossings produce audible joint sounds due to the speed required to perform these motions, and due to the lack of cello sound that, during active performance, conceals background noise. Note that, although all five pieces used for evaluation have left-hand fingerings, these are not used in either the robot or human performances due to potential bias that this may introduce. Left-hand robotic manipulation of the cello is a future application of our work, whereas the current focus is on bowing motions.  
\begin{figure}[h]
    \centering
    \includegraphics[width=1.0\linewidth]{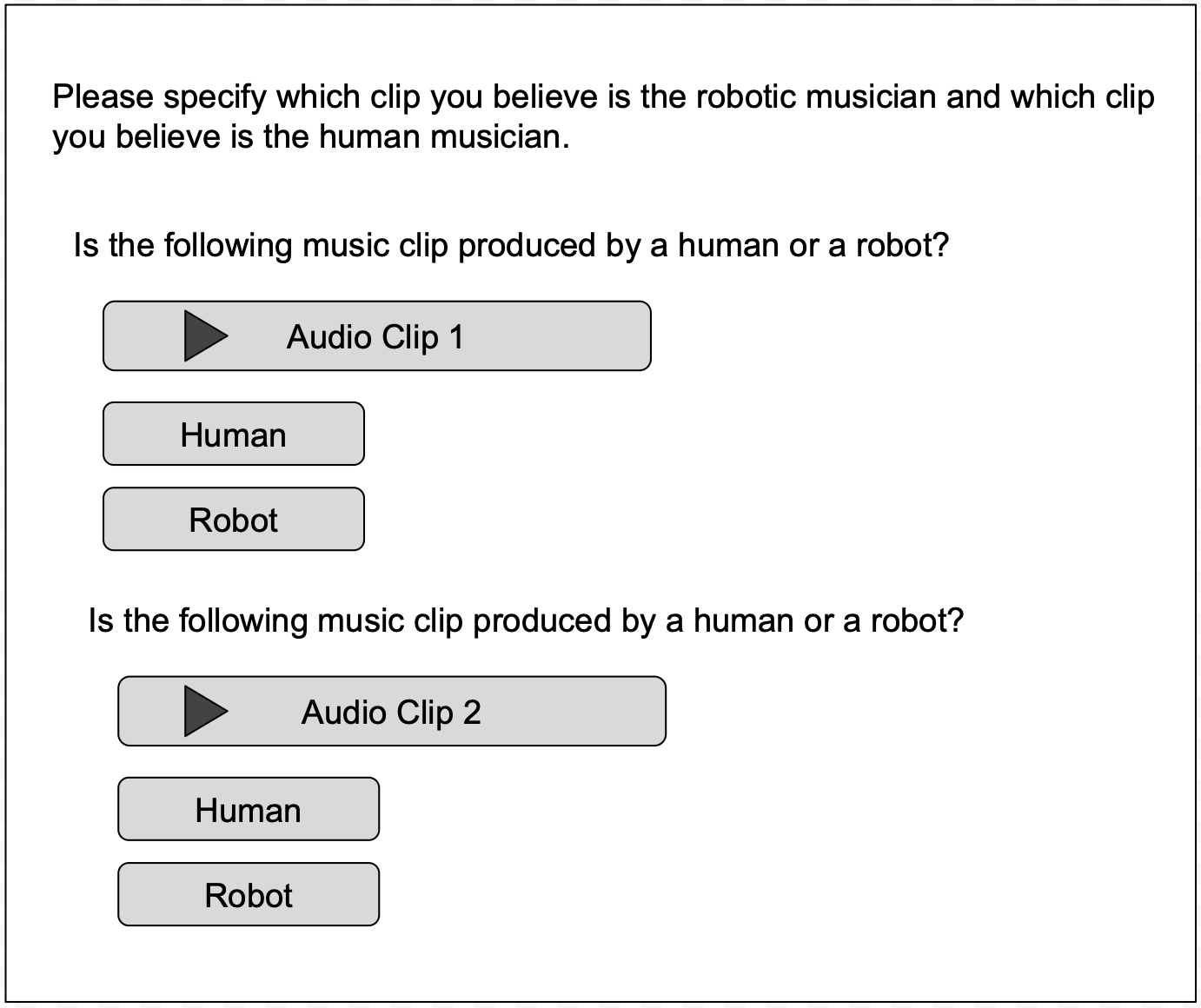}
    \caption{Question for human listeners. The two audio clips are shown in random order, where one unlabeled clip is the robot audio and the other clip is the human audio of the same segment from the musical piece.}
    \label{figure:example_question}
\end{figure}
\subsection{Questionnaire}
The human-subject evaluation consists of two portions: the musical Turing test portion and the musical quality portion. The focus of the musical Turing test portion is in determining whether our robotic musician is distinguishable from that of a human intermediate cellist. This portion consists of 5 questions, where the user is tasked with listening to two 7-15 second audio clips and determining which, if any, of these clips are produced by our robotic musician. An example question is shown through Figure \ref{figure:example_question}.

The accuracy in this section overall was $72.68\%$ for the total population, with a piece-by-piece breakdown shown in Table \ref{tab:piece-clip-accuracy-combined}. To consider our robotic musician's performance properly indistinguishable from human performance, we target an accuracy of $50\%$, as there are two answer choices per question and this is expected accuracy of random guess. Human accuracy from the total participant population is much higher than $50\%$, and as such it suggests that our system can be improved to better replicate a human musician. However, the $27.32\%$ inaccuracy rate also implies that our results show promise, especially when considering future research directions involving imitation learning and use of force sensor data. 

\begin{table}[ht]
\centering
\caption{Accuracy by Musical Piece: Non-Musicians and Total Population}
\label{tab:piece-clip-accuracy-combined}
\setlength{\tabcolsep}{3pt} 
\renewcommand{\arraystretch}{1.15}
\begin{tabular}{p{2.3cm} p{1.2cm} p{2.3cm} p{2.3cm}} 
\toprule
\textbf{Piece} & \textbf{Expected} &
{\centering \textbf{Non-Musicians\\(Correct/Total, \%)}\par} &
{\centering \textbf{Total Population\\(Correct/Total, \%)}\par} \\
\midrule
Allegro           & Robot & 19/32 (59.38) & 94/134 (70.15) \\
                  & Human & 16/32 (50.00) & 84/134 (62.69) \\
Perpetual Motion  & Human & 23/32 (71.88) & 107/133 (80.45) \\
                  & Robot & 22/32 (68.75) & 107/134 (79.85) \\
Twinkle, Twinkle  & Human & 18/31 (58.06) & 110/133 (82.71) \\
                  & Robot & 21/31 (67.74) & 108/132 (81.82) \\
Long, Long Ago    & Robot & 8/29 (27.59)  & 74/129 (57.36) \\
                  & Human & 13/29 (44.83) & 84/129 (65.12) \\
Minuet No. 2      & Robot & 15/29 (51.72) & 92/128 (71.88) \\
                  & Human & 16/29 (55.17) & 95/128 (74.22) \\
\bottomrule
\end{tabular}
\end{table}

Although the total results over the full population of participants showed high accuracy in determining the robotic performance versus the human performance, it is interesting to note that participant accuracy is reduced drastically when only considering non-musicians. When considering this case, overall accuracy drops to $55.88\%$, a significant improvement. The piece-by-piece results for non-musicians can be found in Table \ref{tab:piece-clip-accuracy-combined}. As our survey contains majority ($78.69\%$) musicians, this does not accurately represent the $\sim27\%$ minority of Americans aged 18-24 who play an instrument ~\cite{namm-2003-american-attitudes}. Therefore, a more accurate reflection of robot versus human performance may be found through a less biased population pool. 

Robotic performance for "Perpetual Motion" was significantly weaker than  other pieces used in the study. Best results were obtained with "Long, Long Ago" for both the total and non-musician population. This is likely due to the slow nature of the piece and lack of string crossings, as our robotic musician struggles to play string crossings quickly and produce a clear sound with shorter notes. With that being said, performance in "Allegro", another fast-paced piece, was second-highest overall, showing that our robotic musician is capable of handling fast pieces as long as they have limited string crossings. The main challenge faced with the bowing task of our robot is in creating bowing motions that are both collision-free and efficient in timing. For the sake of successful transitions to all four cello strings, collision avoidance was emphasized over speed in this work. Weakest results were found in "Perpetual Motion" and "Twinkle, Twinkle Little Star". "Perpetual Motion" proves to be an especially difficult challenge for our robotic musician, as there is a high number of string crossings (22 total) as well as short and fast note sequences. Performance in "Twinkle, Twinkle Little Star" is also likely due to the number of fast string crossings required to properly perform the piece.

\begin{figure}[H]
    \centering
    \includegraphics[width=1.0\linewidth]{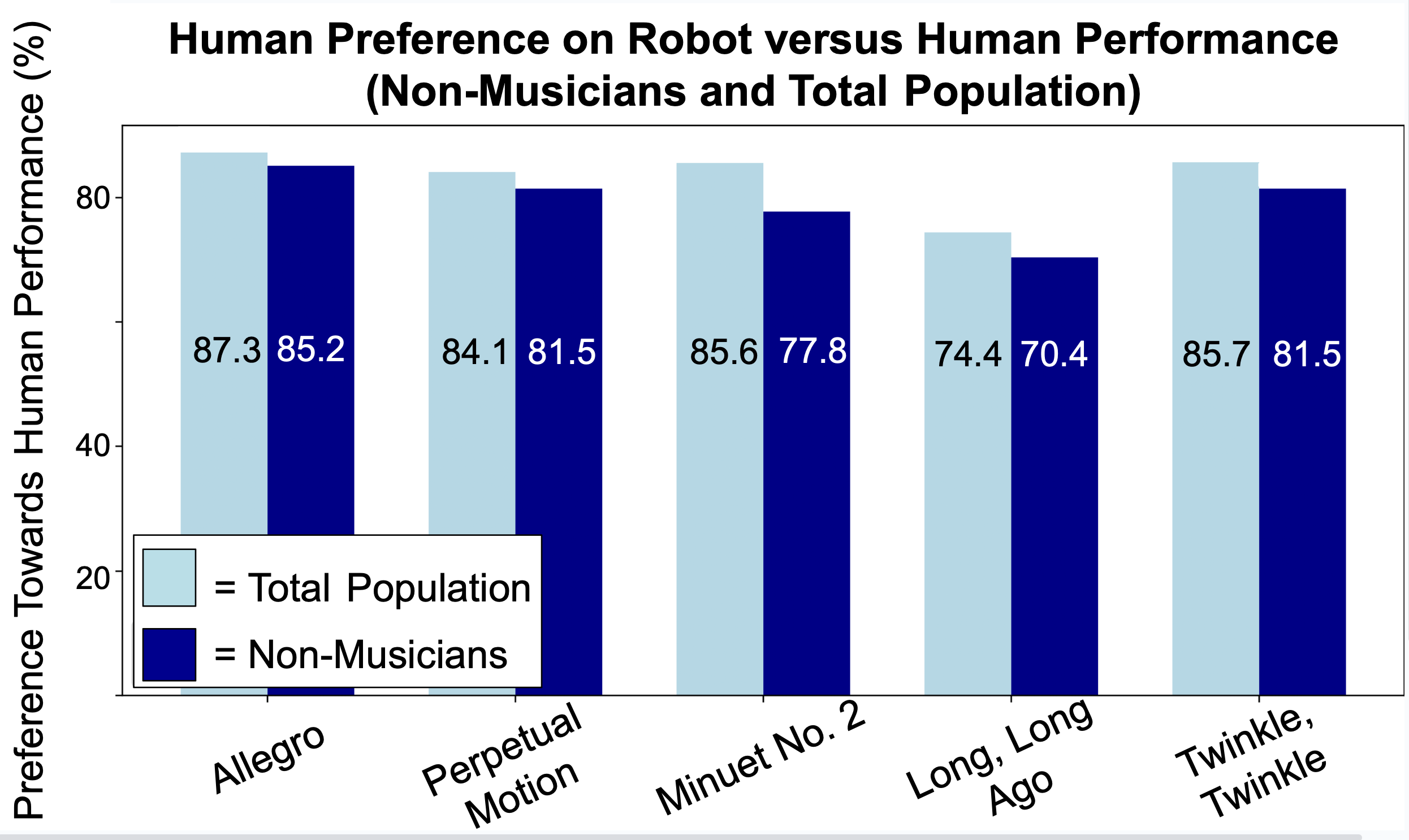}
    \caption{Percentage of participants who prefer human-generated audio (total population and non-musician population).}
    \label{figure:music-pref}
\end{figure}

The Musical Quality and Preferences portion of the study measures: 1) whether the robot-recorded audio is preferred to the human audio by listeners, 2) whether the perceived musical quality of the robotic audio is greater than that of the human audio, and 3) whether there is inherent bias in perception of musical quality by knowing that a musical clip is produced by a robotic performer. To explore the first question of whether the robot-recorded audio is preferred to the human audio, short clips from the five standard pieces used in the study are provided to the participants. They must select which unlabeled audio clip they prefer from each piece, where one provided clip is a robotic recording and the other provided clip is a human recording. From this question, the results are as shown in Figure \ref{figure:music-pref}. For all five pieces, the preference for human-recorded audio is very high, especially with a majority-musician group of participants. However, for "Long, Long Ago", a significant minority of $25.6\%$ of the total population, $29.6\%$ of the non-musician population preferred the robot-generated audio clip. For all other pieces, the preference for robot-generated audio fell between 12.7 - 22.2 \%. These results align with the Musical Turing Test results; Pieces where the robotic performance is hard to distinguish from the human performance are the most preferred robotic performances. 

To assess musical quality of human-performed versus robot-performed audio clips, a series of five questions was given to the participants in the study. For each question, the participant is told at random that either: a) the given audio clip is robot-generated or b) the given audio clip is human-generated. However, the true label of the musical clip may not be the label told to the user. This method seeks to account for potential biases in preference for human-generated audio. It was found that, for a given piece, being told that the audio was human-generated increased quality rating on a scale from 1-5 by 0.47 points on average per-piece, with the largest increase in ranking found for "Long Long Ago" (over half a point increase). This brings out an interesting result that participants, when given the same audio clip, will still prefer the clip which they percieve to be generated by a human. These results can be summarized in table \ref{tab:quality_means_by_piece}. For robot-generated audio clips, the average quality score was 2.614, including both true and false-labeled values. For human-generated audio clips, the average quality score was 3.732. Therefore, there is a significantly higher average quality ranking given to the human-generated audio clips, with a 1.118 point advantage. However, the actual difference in quality score may be lower due to biases in preference for robot-generated audio. 

\begin{table}
\caption{Average quality scores by piece, told label, and actual label}
\label{tab:quality_means_by_piece}
\begin{tabular}{llrr}
\toprule
Piece & Actual Label & Robot Mean & Human Mean \\
\midrule
Allegro & Robot & 2.407 & 2.923 \\
Perpetual Motion & Robot & 1.979 & 2.469 \\
Long, Long Ago & Human & 3.545 & 4.200 \\
Minuet No. 2 & Human & 3.403 & 3.780 \\
Twinkle, Twinkle & Robot & 2.787 & 3.120 \\
\bottomrule
\end{tabular}
\end{table}

\section{DISCUSSION}

\subsection{Limitations and Future Directions}
The main limitations of this work include lack of auditory feedback as well as lack of sensors in the physical robotic environment. As cello performance requires precise positioning and touch sensing, use of force sensing such as through strain gauges would provide a useful extension to the work. For example, the Hyperbow is able to capture many intricies of bowing technique ~\cite{young-hyperbow-2002}. Inclusion of auditory feedback through use of microphones, as well as visual feedback through cameras for positional analysis will be another future step in this project. Specifically, the lack of proper sensors in the robot environment had a negative effect on our goals of utilizing reinforcement learning and behavioral cloning as techniques for improving robotic performance. Proper string contact point, perpendicularity to the string, and bow-force are all required to produce high quality sound. Additionally, sound is challenging to replicate in simulation, and as such our current MuJoCo learning environment is not suitable to achieve high level results with our learning goals. 

Future directions include modifying both the physical and simulation environment of our robotic arm, such that we can obtain more detailed feedback to improve performance and inject higher levels of musical expression. Future systems will also incorporate a robotic fingering mechanism for the left hand, such that performance is not limited to open-string recordings. Lastly, string crossings will be smoothed and improved via learning mechanisms such as reinforcement learning, producing a higher quality and more difficult to distinguish performance from the human listener. 

\subsection{Applications}

This system opens possibilities in several domains. As an educational tool, a robotic cellist can perform bowing motions while the student focuses on the left hand fingerings. Additionally, for people who cannot otherwise play the cello, the robotic cellist may allow them to interact with music in novel ways. Within the performance context, the robotic performer can play alongside human performers, enabling new forms of artistic expression. From a research standpoint, the release of our dataset containing audio, video, and motion recordings provides a benchmark resource for studies in music information retrieval, acoustics, and human–robot interaction. Finally, the system contributes to the broader field of robotic musicianship by demonstrating a framework for end-to-end score-to-motion translation in complex acoustic instruments.

\section{CONCLUSION}

This work introduces an end-to-end pipeline which converts MIDI scores into robotic bowing motions for a full-size cello. Through extensive data collection and evaluation, including a Musical Turing Test study with 132 total participants, we demonstrate that non-musicians are unable to reliably distinguish between robotic and human performances, with average accuracy only $5.88\%$ above chance. These findings highlight that our system can generate convincing performances while generalizing to unseen pieces, establishing a reproducible framework and dataset for evaluating robotic musicianship through human perception. 

\section*{Acknowledgments}

The authors acknowledge the use of AI tools (ChatGPT) for assistance in generating code snippets, refining research ideas, and generating figures. All content was reviewed and finalized by the authors.

\addtolength{\textheight}{-12cm}   




\bibliographystyle{IEEEtran}
\bibliography{references}

\end{document}